\documentclass[runningheads]{llncs}
\usepackage[T1]{fontenc}
\usepackage{subfigure}
\usepackage{graphicx}
\usepackage{amsmath}
\usepackage{url}
\begin{document}
\title{Motion Generation Review: Exploring Deep Learning for Lifelike Animation with Manifold}
\titlerunning{Exploring Deep Learning for Lifelike Animation with Manifold}
%

\author{Jiayi Zhao \inst{1}\orcidID{0009-0007-3789-000X},
 Dongdong Weng \inst{1,2}\orcidID{0000-0003-2352-0896},
 Qiuxin Du \inst{1}\orcidID{0009-0000-5991-9482},
 Zeyu Tian \inst{1}
}
\authorrunning{Zhao. et al.}
%
\institute{Beijing Engineering Research Center of Mixed Reality and Advanced Display, School of Optics and Photonics, Beijing Institute of Technology, China,
\email{\{3120230529@bit.edu.cn, crgj@bit.edu.cn, qiuxin807521@163.com, tianty97@163.com\}}\\
\and Zhengzhou Academy Of Intelligent Technology, Beijing Institute of Technology, China,\\
\email{\{crgj@bit.edu.cn\}}
}

\maketitle              

\begin{abstract}
Human motion generation involves creating natural sequences of human body poses, widely used in gaming, virtual reality, and human-computer interaction. It aims to produce lifelike virtual characters with realistic movements, enhancing virtual agents and immersive experiences. While previous work has focused on motion generation based on signals like movement, music, text, or scene background, the complexity of human motion and its relationships with these signals often results in unsatisfactory outputs. Manifold learning offers a solution by reducing data dimensionality and capturing subspaces of effective motion. In this review, we present a comprehensive overview of manifold applications in human motion generation—one of the first in this domain. We explore methods for extracting manifolds from unstructured data, their application in motion generation, and discuss their advantages and future directions. This survey aims to provide a broad perspective on the field and stimulate new approaches to ongoing challenges.

\end{abstract}

\begin{figure}[ht] 
    \centerline{
    \includegraphics[width=1.3\textwidth]{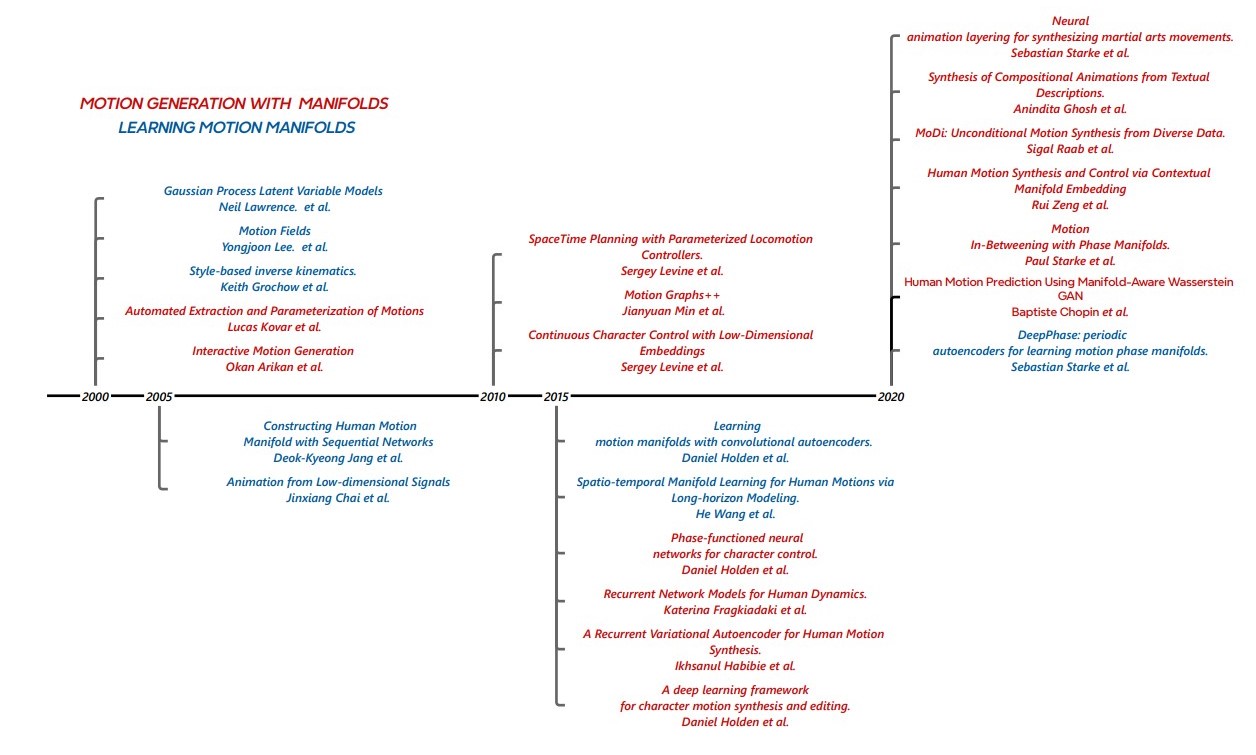}}
    \caption{The typical concept of learning manifold from motion and motion generating with manifold has been proposed in the recent twenty years.}
    \label{cover}
\end{figure}

\section{INTRODUCTION}
With the progression of computer technology, an increasing number of applications and media, such as virtual reality (VR), gaming, and film, are utilizing highly detailed digital human and animal characters. Films like ``Avatar,'' ``The Avengers,'' ``The Lord of the Rings,'' and ``Alita: Battle Angel,'' alongside the booming gaming industry, have showcased the potential of digital characters engaging in complex, lifelike motions. These interactive experiences have elevated the demand for generating high-fidelity and realistic virtual characters, crucial for immersive environments and interactive simulations.In particular, VR environments rely heavily on the seamless integration of virtual characters whose motions mimic the fluidity and naturalness of real-world movements. This not only enhances the immersion but also elevates user interaction in these virtual worlds. Traditionally, Motion Capture (MOCAP) has been the dominant method for producing high-quality motion sequences.\cite{zuo2024loose,liu2011markerless} However, MOCAP technologies are expensive and require specialized equipment, and they are restricted to capturing existing motions, offering limited flexibility when creating new or adaptive movements in real-time environments such as VR or interactive simulations.\cite{moeslund2006survey,gall2009motion} In contrast, data-driven motion generation has gained momentum due to its flexibility, scalability, and reduced need for extensive motion capture setups. Unlike MOCAP, data-driven methods allow for generating novel and diverse movements with minimal input data, making them particularly suitable for real-time applications in immersive technologies like VR, where adaptability and automation are paramount. This method also integrates smoothly with traditional keyframe animation, making it a go-to solution in both film production and gaming environments.

Various techniques have been developed for data-driven motion generation, including Generative Adversarial Networks (GANs)\cite{goodfellow_generative_2014}, Variational Autoencoders (VAEs)\cite{kingma_auto-encoding_2013}, Motion Graphs\cite{arikan_interactive_2002,lee_interactive_2002,kovar_motion_2002}, Reinforcement Learning\cite{sutton1998reinforcement}, and Diffusion Models\cite{ho_denoising_2020}. Each of these techniques brings unique advantages. For example, GANs\cite{goodfellow_generative_2014} enable the generation of realistic and reasonable motions by using adversarial processes between a generator and a discriminator, such as PGGAN\cite{karras_progressive_2018}, Style-GAN\cite{karras_analyzing_2020,karras_style-based_2019}, Manifold-Aware Wasserstein GAN\cite{chopin_human_2021}, while VAEs\cite{kingma_auto-encoding_2013} effectively represent motion data by encoding it into lower-dimensional spaces, facilitating easier manipulation of the motion output\cite{sonderby_ladder_2016,noauthor_neural_2017}. Reinforcement learning models such as DeepMimic, AMP\cite{peng_amp_2021}, ASE\cite{peng_ase_2022}, CALM\cite{tessler_calm_2023}, generate physically accurate and controllable motions, which are essential for VR interactions, where user immersion depends on the believability of the character's responses. Additionally, diffusion models\cite{song2020score}, increasingly prominent in generative fields like image and audio synthesis, have recently been applied to motion generation\cite{huang_diffusion-based_2023,shafir_human_2023,tevet_human_2022,zhang_motiondiuse_2024,zhang_remodiffuse_2023}, offering a novel approach to generating natural human motions for immersive environments. Previous research has focused on generating human motion based on fundamental conditional signals such as motion\cite{yu_structure-aware_2020,degardin_generative_2022,lucas_posegpt_2022,petrovich_action-conditioned_2021}, music\cite{le_music-driven_2023,tang_dance_2018,li_ai_2021}, text\cite{ahn_text2action_2017,ahuja_language2pose_2019,petrovich_temos_2022,jiang_motiongpt_2023}, video\cite{zhu_motionbert_2023}, and the scene backgrounds\cite{huang_diffusion-based_2023,zhang_place_2020,hassan_populating_2021}.

Many studies use unstructured motion data from Motion Capture (MOCAP) for generating motion sequences, representing digital character poses. Common databases include CMU\cite{CMUMocap}, Human3.6\cite{ionescu2013human3}, NTU RGB+D\cite{liu2019ntu}, and HumanML3D\cite{guo2022generating}, with data typically represented by joint angles and positions. However, challenges remain. First, validating the realism of generated motions is difficult due to the abstract and complex nature of motion sequences. Ensuring motions are realistic and executable by humans or quadrupeds is especially challenging for VR and immersive applications. Second, many methods require extensive manual pre-processing like segmentation, alignment, and labeling, where errors can lead to animation failures. This complicates full automation and often demands technical expertise.

\begin{figure}[t]
    \centering
    \subfigure[]{%
        \includegraphics[width=0.25\textwidth]{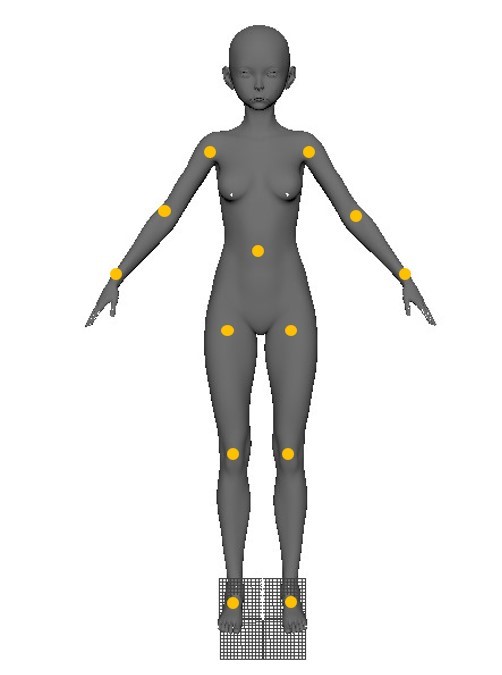}
        \label{2}
    }
    \subfigure[]{%
        \includegraphics[width=0.22\textwidth]{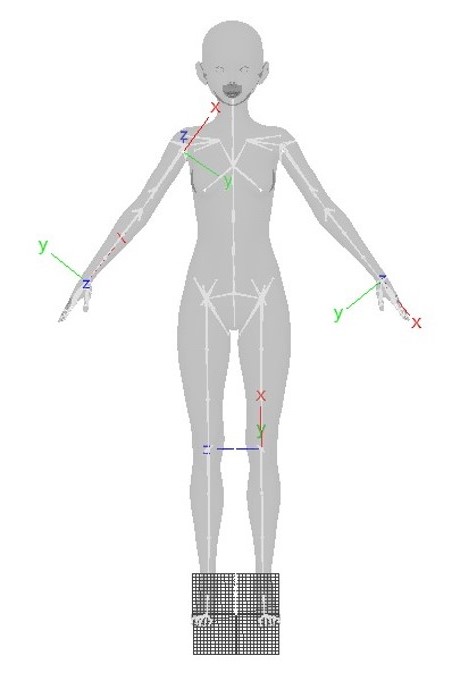}
        \label{3}
    }
    \subfigure[]{%
        \includegraphics[width=0.35\textwidth]{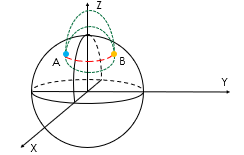}
        \label{4}
    }
    \caption{Typical human pose and shape representations in (a) keypoint-based and (b) rotation-based (c) Find a valid motion in the entire motion space. A point and B point represent two statuses, there are many different motions between A and B like green lines, only the motion on the surface is valid like the red line.}
\end{figure}

Learning a manifold from MOCAP data and performing motion generation in this subspace can facilitate the discrimination between effective and ineffective motions, leading to enhanced automation. This is because this manifold is learning from the motion capture actor, which means constructing a valid motion subspace within the entire motion space. At the same time, this process is a kind of data dimensionality reduction that can reduce the required computing power. In immersive environments such as VR, where real-time performance is critical, reducing computational costs while maintaining motion quality is a significant advantage. Therefore, If we find good techniques for learning manifolds, it’s easy for valid motion generating, interpolating, etc.

In this research review, we introduced the manifold in motion generation. And summarized methods for obtaining motion manifolds and the application of manifolds in motion generation, as shown in Figure \ref{cover}. 

\section{PRELIMINARIES}

\subsection{Motion Data Representation}
Motion data representations are generally categorized as keypoint-based or rotation-based\cite{zhu_human_2023}, and they can be converted between forms. Keypoint-based data represent the human body using specific kinematic points(e.g., joints)in 2D or 3D space, as shown in Figure \ref{2}. This data can be easily obtained from motion capture systems but is less suitable for animation or robotics due to difficulty in capturing relationships between keypoints. Each frame of a motion sequence records the position of these keypoints. Rotation-based data represent the human body through joint rotation angles in a hierarchical structure, as shown in Figure \ref{3}. Formats for these angles include quaternions, euler angles, rotation vectors, and RT matrices. Certainly, there are other representations of human body posture\cite{ma20233d,zhang2021we}. 

The Skinned Multi-Person Linear (SMPL) model\cite{loper_smpl_2023} is a popular parametric 3D human model used in pose estimation, motion capture, and virtual character modeling. It uses Linear Blend Skinning (LBS) to realistically depict various poses and body shapes. Extensions of SMPL include SMPL-X\cite{pavlakos_expressive_2019}, GHUM\cite{xu_ghum_2020}, and STAR\cite{osman_star_2020}.

\subsection{Manifolds}
Motion manifold is a new representation of animation data. In this section we introduce the definition of motion manifolds, the reason they can be used in the motion of digital humans and a kind of motion manifold like phase Manifold.

\textbf{Motion Manifold}
Strictly, manifold is a concept in mathematics and computer science that refers to a space that locally resembles Euclidean space.\cite{lee2022manifolds,bishop2011geometry}  Manifold has a local space that is homomorphic to Euclidean space. The meaning of it is the local space has characteristics of Euclidean space that is the distance in this local space equivalent to Euclidean distance.\cite{tu2011manifolds}

Motion manifold is a low-dimensional geometric structure used to represent and analyze motion data.\cite{bishop2011geometry} By embedding motion samples into a continuous manifold, it captures the intrinsic features and variations of motion. Each point on the manifold corresponds to a specific pose or motion state, and the geometric properties of the manifold make interpolation and generation between different poses feasible. Therefore, by performing interpolation or other operations on this manifold, new and smooth motion trajectories can be generated.

Compared to traditional motion sequences, the motion manifold is generally considered low-dimensional. Traditional motion sequences may contain a large amount of redundant data, while the motion manifold focuses on capturing the essential features of motion, eliminating unnecessary details and thus reducing dimensionality. Motion data often exhibits inherent geometric structures. Manifold learning can effectively identify and represent this structure, enabling motion state variations to be expressed in fewer dimensions.In summary, the motion manifold removes redundancies and emphasizes key features, allowing high-dimensional motion data to be efficiently represented as a low-dimensional manifold, thereby enhancing the efficiency of processing and generating motions. Thus, it provides us with a dimensionality reduction method.\cite{holden_reducing_2017} 

Animation data is typically represented as a time series of joint angles and positions, where each frame consists of high-dimensional vectors means the character's pose. This representation is suitable for data processing and computing. However, using this representation to compute may obtain unnatural results like the large differences between frames to make joints move too fast or unreasonable joint angles or positions resulting in physically impossible motion. Because all bones of the body are highly correlated in a motion, which should satisfy the constraint of bone lengths and joint rotations. 

In space of motion, define the motion manifold as a subspace of valid motion.\cite{holden_reducing_2017} It is of great interest for researchers to find motion manifolds in the whole space. In this subspace, it is very convenient to find a reasonable motion between two statuses. If we consider the entire three-dimensional space as motion space. Motion manifold is the surface of the ball in the space, as shown in Figure \ref{4}. On the ball, A point and B point represent two digital human statuses. There are many different motions between A and B. Only the motion on the surface is valid. Other ways are not reasonable motions. Therefore, the motion manifold is a kind of dimensionality reduction, it describes the valid motion in whole space. It can provide more natural and smooth motion. At the same time, it still allows to reproduce complex movements of the human body.

\textbf{Motion Phase Manifold}
The motion manifold does not have a fixed data structure.\cite{lee2022manifolds} Motion phase manifold is a type of motion manifold.\cite{holden_deep_2016} About the data-driven method, MOCAP data of a single type of locomotion are dense, while transitions between different types of motions are very sparse. Thus, it causes a problem, which is modeling or learning the transitions between different motions is difficult, which requires attention to the phase of motion, and finding relationships of different types of motions in phase space. The phase manifold describes the manifold of the motion phase like fight, the position of the hands and legs change over time, and this alteration is the phase, like Figure \ref{3}. Phase manifold refers to the manifold that describes alterations of phase. The features in the phase manifold describe the frame timing of the motion. It is great for aligning motions within the same category or between motions of different categories, whose meaning is the phase manifold can be an effective input feature for motion synthesis or motion matching. In the article of Holden.\cite{holden_deep_2016} delimit phase manifold can be computed by each frame of the motion sequence.
\begin{equation}
P_{2i-1}^{(t)} = A_i^{(t)}*\sin (2\pi * S_i^{(t)}),\label{eq:1}
\end{equation}
\begin{equation}
P_{2i}^{(t)} = A_i^{(t)}*\cos(2\pi * S_i^{(t)}),\label{eq:2}      
\end{equation}
\begin{equation}
(s_x , s_y) = FC (L_i ), S_i = atan2(s_x , s_y),\label{eq:3}      
\end{equation}
\begin{equation}
S_i = atan2(s_x , s_y),\label{eq:4}     
\end{equation}

where A is amplitude, F is frequency, B is offset, S is phase shift and $S \in R^{M}$. They are the periodic parameters. FC is a fully-connected layer. The phase features of phase should cluster the animations in both space and time, like Eq. \ref{eq:1}-\ref{eq:4}. Features such as frequency and amplitude remain near-constant over time and do not help well for alignment purposes. After modeling the transition between motion and phase, which helps us to get smooth and coherent conversion between different motions. Thereby, we can obtain copious motions.

\section{LEARNING MOTION MANIFOLDS}
Manifold is a general term for general geometric objects, including curves and surfaces of various dimensions. Manifold learning is a type of dimensionality reduction method that re-represents a set of data in a high-dimensional space in a low-dimensional space. In motion, the motion manifold generalizes a high-dimensional vector to describe poses of character.   

Generally, the original motion data we can get is from MOCAP. Thus, generating motion manifolds from human motion data is a great interest in machine learning and computer graphics. Firstly, use the principal component analysis (PCA) to get motion manifold from the original motion data. PCA is effective in modeling the manifold of human gait. Chai.\cite{chai_performance_2005} uses local PCA to obtain motion manifolds of various types of human motion data, which is used for motion synthesis.  Lawrence.\cite{lawrence_gaussian_2003} introduces a new underlying probabilistic model for PCA, which is a particular Gaussian process before a mapping from a latent space to the observed data-space (GPLVM).  And Grochow.\cite{grochow_style-based_2004} use GPLVM for motion synthesis. However, PCA often has a certain degree of ambiguity and is not as complete as the original data. In another way, Lee.\cite{lee_motion_2014} proposed a data structure that is motion fields. In this data structure, the user can effectively and interactively control the characters by using motion manifold. Although these methods can get manifold, they need a lot of pre-processing, including computing the distance of each frame. However, the shortcoming of these methods is only encode the temporal aspect of motion. They do not use deep learning that is unable to obtain manifolds for large amounts of data. Moreover, they exhibit poor robustness to different types of motions.

With the advancement of computing power and deep learning, more methods now leverage deep learning techniques. Holden et al.\cite{holden_learning_2015} proposed using Convolutional Autoencoders to learn a manifold of human motion data from time-series inputs. This method, which uses a three-layer convolutional autoencoder, is simple yet effective. Once the motion manifold is obtained, it can be applied to tasks such as fixing corrupt data, filling missing data, motion in-betweening, and comparison. It offers better scalability and handles large datasets more effectively than traditional methods, outperforming PCA-like techniques in non-linear tasks.

Jang et al.\cite{jang_constructing_2020} proposed a recurrent neural network-based method to construct a motion manifold capable of representing long sequences of diverse human motions. They introduced new regularization terms for the manifold and used two complementary decoders to predict joint rotations and velocities. A forward kinematics layer was added to account for both joint rotation and position errors. Their method uses an end-to-end unsupervised approach, minimizing the difference between the ground truth and reconstructed motion space distributions. The network consists of an RNN-based encoder with Gated Recurrent Units (GRU), a regularizer to align the encoded motion distribution with a prior distribution, and two decoders to map the latent space to joint angles and velocities. 

Unlike images, human motion is represented on a graph structure, and the lack of spatial feature extraction can lead to convergence to a ``mean posture,'' where spatial variation in movement disappears. Temporally, motion exhibits multi-modality and dynamic variations—different motions can share prefixes or endings, and identical postures may vary in timing. To address this, Wang et al.\cite{wang_spatio-temporal_2019} proposed the Spatio-temporal Recurrent Neural Network (STRNN), which models both spatial and temporal variances to avoid generating ``mean postures.'' STRNN comprises three sub-networks: a spatial network for hierarchical modeling of body parts, a temporal network for learning dependencies in long motion sequences, and a residual network to filter out high-frequency noise.

For phase manifold, Starke.\cite{starke_deepphase_2022} proposed a novel neural network architecture for character motion synthesis called the Periodic Autoencoder, which can learn periodic features from large unstructured motion datasets like phase. This method extracts multi-dimensional phases from whole body data, which validly clusters all kinds of motion animations and produces a phase manifold of them. In this measure, use convolutional autoencoders to transform the motion space into a phase manifold, whose structure is similar to Holden.\cite{holden_learning_2015} 
The periodic parameters for unstructured motion data can be computed for each frame by shifting the periodic autoencoder along the motion curves, like Eq. \ref{eq:1}-\ref{eq:2}. After computing the phase manifold, it can be used to do many things like motion control, motion matching, stylistic movements, dance motion synthesis, etc.


\section{MANIFOLDS IN MOTION GENERATION}
\subsection{Motion Synthesis}
Motion synthesis is a task using current MOCAP data to produce a series of new motions like blending motion of the same class to create new motion. For data-driven character motion synthesis, there are two ways to achieve, physical-based and data-based. Generally, approaches to motion synthesis are kernel-based methods, which synthesize motions via blending MOCAP data, and interactive character control, where user instructions to synthesize motions using a motion database. and deep learning.

\textbf{Kernel-based methods}, Radial basis functions (RBF) is effective for blending motions of the same class. Rose.\cite{rose_verbs_1998} defines motions of the same class as ‘verbs’ and applies RBF to motion interpolation based on the direction the character moves toward. To automate motion synthesis, Kovar.\cite{kovar_automated_2004} computes the similarity of the movements and aligns them via dynamic time warping. But, RBF is dependent on data, if noise and variance of data are too many, it easily overfits the data. Therefore, Grochow.\cite{grochow_style-based_2004} uses GPLVM to map the motion to low-dimensional space to intuitively control the characters. Levine.\cite{levine_continuous_2012}apply reinforcement learning to compute the optimal motion in the low-dimensional space, dimensionality reduced by GPLVM.  Moreover, Kernel-based methods suffer from large memory costs, which cause motion can be blended is limited.

\textbf{Interactive character control}, a mechanism is applied to produce a series of consecutive motions via user-provided high-level commands. Motion graph\cite{min_motion_2012} is an effective data structure for this purpose. Owing to Motion graphs are automatically computed from a large MOCAP data and only replay the MOCAP data. Arikan.\cite{arikan_interactive_2002} propose a technique to blend motion data of the same class to enrich the dataset. For automating this process and finding optimum motion, reinforcement learning is applied to motion synthesis.\cite{levine_space-time_2011} However, reinforcement learning needs an amount of pre-computation, which will Increase exponentially with the number of characters. Thus, reinforcement learning requires a method to reduce the number of states and data dimensionality.

\textbf{Deep learning methods}, a technology that obtains consecutive and reasonable motions via applying deep neural networks to Learn unstructured motion data to produce new motion. Holden.\cite{holden_learning_2015} apply a convolutional autoencoder to the CMU MOCAP database and show the learned representation achieves good performance. Baptiste Chopin\cite{chopin_human_2021} employed a concise manifold-valued representation of human motion to address both the predicted motion's discontinuity and the performance decline over extended time horizons.

Most previous data-driven methods require extensive manual preprocessing, such as aligning, separating, and labeling motion data. Mistakes in preprocessing can lead to unreasonable results, making full automation challenging. To address this, Holden et al.\cite{holden_deep_2016} introduced a framework for synthesizing character movements based on high-level parameters instead of low-level details. This approach employs a convolutional autoencoder to learn the mapping between motion and the manifold represented by the network's hidden units, similar to Holden et al. \cite{holden_learning_2015}. High-level parameters, such as the root trajectory projected onto terrain or the end effector trajectories of hands and feet, are used as inputs. A feedforward network maps these parameters to the motion manifold, integrating current motion manifolds to create new ones. To ensure controllability, generated motions can be edited through optimization within the motion manifold space using an editing network, which maintains the naturalness and smoothness of the motion. 

Previous approaches, apply one-dimensional convolutions along the timeline\cite{holden_deep_2016} or rely on Recurrent Neural Networks (RNN) to implement time series models with feedback structures.\cite{fragkiadaki_recurrent_2015} However, these methods are difficult to predict frames of motion more than a little bit in the future, resulting in converging to a static pose. Therefore, Habibie.\cite{habibie_recurrent_2017} proposes a novel approach to motion synthesis, which can accept user-provided control signals and encode the signals into a variational inference framework that learns the manifold of human motion. To predict the far future of the motion, they combine variational inference, the consideration of a control signal and several deep learning modules to produce high-quality long-term motion, which is VAE-LSTM architecture. This system is composed of an encoder and an autoregressive decoder. For the encoding, they encode the control signals and joint positions as manifolds, processing in manifolds.

To synthesize digital human motion from a given data distribution is a challenging task, especially when the dataset is highly diverse, unstructured, and unlabeled. And related works need priori conditions like root trajectory, trajectories of the end effectors\cite{holden_deep_2016}, or focus on synthesizing specific types of motion with limited diversity\cite{holden_phase-functioned_2017}. Raab.\cite{raab_modi_2023} presents MoDi, an unconditional generative model that synthesizes diverse motions. The core of this method is a deep generative model trained in an unsupervised manner on an extremely diverse, unstructured and unlabeled motion dataset. Because many tasks in the motion domain are ill-posed. To be suitable for different tasks, present an encoder that inverts real motions into MoDi’s natural motion manifold.

In addition to using the network to map the motion into the manifold, there is another way that is embedding the Gaussian Process Latent Variable Model(GPLVM) in deep learning framework to precisely model motion dynamics and rapid control\cite{zeng_human_2021}. Their model is encoder-decoder form. Their workflow is in two stages. Firstly, apply GPLVM to project motion poses to manifold, where motion states could be clearly recognized. Secondly, the Recurrent Neural Network (RNN) encoder makes temporal latent prediction via the manifold of previous motion and control states. An attention module is used to morph the prediction by calculating similarities between the control states and predicted states. In the end, the GP decoder reconstructs motion states back to motion frames. In this way, no need to train the encoder on the manifold in advance. GPLVM is a nonlinear, nonparametric probabilistic model that can better handle the randomness and uncertainty of motion.

Manifolds not only apply to ordinary motion blends but also used to achieve text-to-motion. Ghosh.\cite{ghosh_synthesis_2005} presents a new technique, a hierarchical two-stream model, for generating compositional motions, which can handle complex sentences as the input. The input sentence can not only describe a simple action but also can describe a human performing multiple actions either sequentially or simultaneously, like “a person is stretching his arms, taking them down, walking forwards for four steps and raising them again”. the hierarchical two-stream model consists of a pose encoder, sentence encoder and pose decoder. The workflow is three steps. Firstly, use the pose encoder to hierarchical model motion, get features based on five major parts of the body and combine those features hierarchically. Secondly, use sentence encoder to represent the text, apply BERT as a contextualized language model, and at the same time apply Long-Short Term Memory units (LSTM) to capture the long-range dependencies of complex sentences. Besides, the motion modeled by the pose encoder and described by input text will map into manifold after processing. Finally, the pose decoder constructs the output motion and calculates loss to train the network.

In motion synthesis, traditional methods often rely on rule-driven or model-based algorithms, which, while capable of generating reasonable motions in specific scenarios, struggle to capture complex patterns in high-dimensional spaces. In contrast, motion manifolds embed high-dimensional motion data into a low-dimensional representation, better capturing the underlying structure and generating more natural, realistic movements. This data-driven approach ensures that generated motions maintain temporal and spatial continuity while preserving the characteristics of the original motion. Additionally, motion manifolds effectively mitigate artifacts resulting from insufficient motion modeling in traditional methods. However, their effectiveness depends heavily on the quality and diversity of the training data; insufficient or narrowly focused training data can significantly compromise the quality of the generated motions.

\subsection{Motion Controller}
Motion controller is a real-time data-driven controller for virtual characters that can allow humans via this real-time control of the movement of digital characters. They must be capable of learning from vast amounts of data, should not require extensive manual data pre-processing, and must execute extremely quickly at runtime with low memory requirements. The gap between changes in state and natural randomness becomes challenging to eliminate ambiguity. With the continuous development of deep learning and neural network technologies, there are more and more effective approaches.

Holden et al.\cite{holden_phase-functioned_2017} introduced a real-time character control mechanism using a Phase-Functioned Neural Network that learns motion capture data while considering environmental factors. This architecture allows characters to perform actions like walking, running, jumping, and climbing based on environmental conditions. The system comprises a prediction network that forecasts motions and a phase function that computes network weights each frame. The phase function predicts the next motion type using input parameters and motion data, applying cubic Catmull-Rom splines to assist in action prediction within a manifold space.

A key challenge in computer animation is the interactive synthesis of novel character actions. Starke et al.\cite{starke_neural_2021} proposed a deep learning framework to produce a wide variety of controllable movements from raw motion capture data. The framework consists of three modules: a motion generator, control modules, and a control interface. The motion generator learns the entire motion manifold from unstructured data using an expert-blended network. Control modules create future motion trajectories for various active behaviors, which are then processed through addition, overlay, or blending operations in the control interface before being used to generate novel full-body poses. These control modules can take various forms, including neural networks, physics-based simulations, and user-driven tools, allowing for flexible editing of trajectories. This deep learning framework is adaptable and requires minimal training data.


In motion controllers, motion manifolds improve efficiency and naturalness by simplifying the control of complex motions. Traditional controllers, often rule-based or physics-driven, struggle with flexibility and produce unnatural discontinuities. By mapping high-dimensional motion states into a low-dimensional space, motion manifolds allow for more efficient path planning and smoother transitions. They also enable controllers to handle diverse movements more flexibly. However, controlling within the manifold requires a higher learning demand, as the controller must adapt to the low-dimensional space. Additionally, since the manifold is constructed from existing data, it may struggle with new or unseen motion states, leading to poor performance in extreme or novel cases.

\subsection{Motion in-betweening}


Motion in-betweening uses computer algorithms to generate intermediate frames between keyframes, ensuring smooth transitions in animation. When keyframes have inconsistent durations, systems often rely on linear interpolation, which can produce unrealistic, overly smooth motion. Starke et al.\cite{starke_motion_2023} proposed a novel motion in-betweening framework that learns from a motion capture database using a phase manifold. The system includes a gating network that blends phase segments from a periodic autoencoder and a motion prediction network that takes the current pose, trajectory, contact, and control variables to predict the next frame.

Traditional linear interpolation methods can handle transitions but often result in unnatural artifacts, especially with complex motions. The motion manifold, by operating in a low-dimensional space, enables smoother and more natural transitions while preserving motion consistency and reducing artifacts. However, its effectiveness depends on the quality of training data. For complex or extreme motions not represented in the dataset, the manifold may fail to capture necessary patterns, leading to suboptimal interpolation results.

\section{CONCLUSION AND FUTURE WORK}
Manifolds are widely used in synthesis, control, and motion in-betweening, offering superior performance in generating natural and smooth movements by encapsulating valid human motions. Compared to other methods, manifolds also reduce computational complexity. However, they struggle with interactions involving the environment and other elements because interactive features are not effectively mapped to the motion manifold. Existing methods\cite{holden_deep_2016} handle simple tasks like walking, jumping, and running but fail in complex environments with interactions, such as between multiple characters or with objects. This limitation arises because the interactive dynamics are not incorporated into the manifold, which affects motion generation in more challenging scenes, such as dancing, obstacle navigation, or object manipulation.

In future work, I believe the direction is about how to effectively encode external influencing factors into the manifold space, aiding in the discovery of movements that meet specified requirements. For instance, in text-to-motion scenarios, the challenge lies in encoding the objective constraints described in the text into the manifold. In the case of scene-to-motion scenarios, it involves encoding the objective constraints of the scene into the manifold. In complex interactive environments, the focus is on encoding the motion of other entities and objects into the manifold, ensuring effective influence on the predicted actions and thereby meeting diverse requirements.

\section{ACKNOWLEDGEMENT}
This work was supported by the Strategic research and consulting project of Chinese Academy of Engineering (Grant No. 2023-HY-14).

%
%
\bibliographystyle{splncs04}
\bibliography{reference}

\begin{thebibliography}{10}
\providecommand{\url}[1]{\texttt{#1}}
\providecommand{\urlprefix}{URL }
\providecommand{\doi}[1]{https://doi.org/#1}

\bibitem{ahn_text2action_2017}
Ahn, H., Ha, T., Choi, Y., Yoo, H., Oh, S.: {Text2Action}: {Generative} {Adversarial} {Synthesis} from {Language} to {Action} (Oct 2017), \url{http://arxiv.org/abs/1710.05298}, arXiv:1710.05298 [cs]

\bibitem{ahuja_language2pose_2019}
Ahuja, C., Morency, L.P.: {Language2Pose}: {Natural} {Language} {Grounded} {Pose} {Forecasting} (Nov 2019), \url{http://arxiv.org/abs/1907.01108}, arXiv:1907.01108 [cs]

\bibitem{arikan_interactive_2002}
Arikan, O., Forsyth, D.A.: Interactive motion generation from examples. ACM Trans. Graph.  \textbf{21}(3),  483–490 (Jul 2002). \doi{10.1145/566654.566606}, \url{https://doi.org/10.1145/566654.566606}

\bibitem{bishop2011geometry}
Bishop, R.L., Crittenden, R.J.: Geometry of Manifolds: Geometry of Manifolds. Academic press (2011)

\bibitem{chai_performance_2005}
Chai, J., Hodgins, J.K.: Performance animation from low-dimensional control signals. ACM Trans. Graph.  \textbf{24}(3),  686–696 (Jul 2005). \doi{10.1145/1073204.1073248}, \url{https://doi.org/10.1145/1073204.1073248}

\bibitem{chopin_human_2021}
Chopin, B., Otberdout, N., Daoudi, M., Bartolo, A.: Human {Motion} {Prediction} {Using} {Manifold}-{Aware} {Wasserstein} {GAN} (Jul 2021), \url{http://arxiv.org/abs/2105.08715}, arXiv:2105.08715 [cs]

\bibitem{degardin_generative_2022}
Degardin, B., Neves, J., Lopes, V., Brito, J., Yaghoubi, E., Proen{\c{c}}a, H.: Generative adversarial graph convolutional networks for human action synthesis. In: Proceedings of the IEEE/CVF Winter Conference on Applications of Computer Vision. pp. 1150--1159 (2022)

\bibitem{fragkiadaki_recurrent_2015}
Fragkiadaki, K., Levine, S., Felsen, P., Malik, J.: Recurrent {Network} {Models} for {Human} {Dynamics}. In: 2015 {IEEE} {International} {Conference} on {Computer} {Vision} ({ICCV}). pp. 4346--4354. IEEE, Santiago, Chile (Dec 2015). \doi{10.1109/ICCV.2015.494}, \url{http://ieeexplore.ieee.org/document/7410851/}

\bibitem{gall2009motion}
Gall, J., Stoll, C., De~Aguiar, E., Theobalt, C., Rosenhahn, B., Seidel, H.P.: Motion capture using joint skeleton tracking and surface estimation. In: 2009 IEEE Conference on Computer Vision and Pattern Recognition. pp. 1746--1753. Ieee (2009)

\bibitem{ghosh_synthesis_2005}
Ghosh, A., Cheema, N., Oguz, C., Theobalt, C., Slusallek, P.: Synthesis of compositional animations from textual descriptions. In: Proceedings of the IEEE/CVF international conference on computer vision. pp. 1396--1406 (2021)

\bibitem{goodfellow_generative_2014}
Goodfellow, I., Pouget-Abadie, J., Mirza, M., Xu, B., Warde-Farley, D., Ozair, S., Courville, A., Bengio, Y.: Generative adversarial networks. Commun. ACM  \textbf{63}(11),  139–144 (Oct 2020). \doi{10.1145/3422622}, \url{https://doi.org/10.1145/3422622}

\bibitem{grochow_style-based_2004}
Grochow, K., Martin, S.L., Hertzmann, A., Popović, Z.: Style-based inverse kinematics. In: {ACM} {SIGGRAPH} 2004 {Papers}. pp. 522--531. ACM, Los Angeles California (Aug 2004). \doi{10.1145/1186562.1015755}, \url{https://dl.acm.org/doi/10.1145/1186562.1015755}

\bibitem{guo2022generating}
Guo, C., Zou, S., Zuo, X., Wang, S., Ji, W., Li, X., Cheng, L.: Generating diverse and natural 3d human motions from text. In: Proceedings of the IEEE/CVF Conference on Computer Vision and Pattern Recognition. pp. 5152--5161 (2022)

\bibitem{habibie_recurrent_2017}
Habibie, I., Holden, D., Schwarz, J., Yearsley, J., Komura, T.: A {Recurrent} {Variational} {Autoencoder} for {Human} {Motion} {Synthesis}. In: Procedings of the {British} {Machine} {Vision} {Conference} 2017. p.~119. British Machine Vision Association, London, UK (2017). \doi{10.5244/C.31.119}, \url{http://www.bmva.org/bmvc/2017/papers/paper119/index.html}

\bibitem{hassan_populating_2021}
Hassan, M., Ghosh, P., Tesch, J., Tzionas, D., Black, M.J.: Populating {3D} scenes by learning human-scene interaction. In: Proceedings of the {IEEE}/{CVF} {Conference} on {Computer} {Vision} and {Pattern} {Recognition}. pp. 14708--14718 (2021)

\bibitem{ho_denoising_2020}
Ho, J., Jain, A., Abbeel, P.: Denoising {Diffusion} {Probabilistic} {Models} (Dec 2020), \url{http://arxiv.org/abs/2006.11239}, arXiv:2006.11239 [cs, stat]

\bibitem{CMUMocap}
Hodgins, C.M.U.: Cmu graphics lab motion capture database. \url{http://mocap.cs.cmu.edu/} (2015)

\bibitem{holden_reducing_2017}
Holden, D.: Reducing animator keyframes. Ph.D. thesis, The University of Edinburgh (2017)

\bibitem{holden_phase-functioned_2017}
Holden, D., Komura, T., Saito, J.: Phase-functioned neural networks for character control. ACM Transactions on Graphics  \textbf{36}(4),  1--13 (Aug 2017). \doi{10.1145/3072959.3073663}, \url{https://dl.acm.org/doi/10.1145/3072959.3073663}

\bibitem{holden_deep_2016}
Holden, D., Saito, J., Komura, T.: A deep learning framework for character motion synthesis and editing. ACM Transactions on Graphics  \textbf{35}(4),  1--11 (Jul 2016). \doi{10.1145/2897824.2925975}, \url{https://dl.acm.org/doi/10.1145/2897824.2925975}

\bibitem{holden_learning_2015}
Holden, D., Saito, J., Komura, T., Joyce, T.: Learning motion manifolds with convolutional autoencoders. In: {SIGGRAPH} {Asia} 2015 {Technical} {Briefs}. pp.~1--4. ACM, Kobe Japan (Nov 2015). \doi{10.1145/2820903.2820918}, \url{https://dl.acm.org/doi/10.1145/2820903.2820918}

\bibitem{huang_diffusion-based_2023}
Huang, S., Wang, Z., Li, P., Jia, B., Liu, T., Zhu, Y., Liang, W., Zhu, S.C.: Diffusion-based {Generation}, {Optimization}, and {Planning} in {3D} {Scenes} (Jan 2023), \url{http://arxiv.org/abs/2301.06015}, arXiv:2301.06015 [cs]

\bibitem{ionescu2013human3}
Ionescu, C., Papava, D., Olaru, V., Sminchisescu, C.: Human3. 6m: Large scale datasets and predictive methods for 3d human sensing in natural environments. IEEE transactions on pattern analysis and machine intelligence  \textbf{36}(7),  1325--1339 (2013)

\bibitem{jang_constructing_2020}
Jang, D.K., Lee, S.H.: Constructing {Human} {Motion} {Manifold} with {Sequential} {Networks}. Computer Graphics Forum  \textbf{39}(6),  314--324 (Sep 2020). \doi{10.1111/cgf.14028}, \url{http://arxiv.org/abs/2005.14370}, arXiv:2005.14370 [cs]

\bibitem{jiang_motiongpt_2023}
Jiang, B., Chen, X., Liu, W., Yu, J., Yu, G., Chen, T.: {MotionGPT}: {Human} {Motion} as a {Foreign} {Language} (Jul 2023), \url{http://arxiv.org/abs/2306.14795}, arXiv:2306.14795 [cs]

\bibitem{karras_progressive_2018}
Karras, T., Aila, T., Laine, S., Lehtinen, J.: Progressive {Growing} of {GANs} for {Improved} {Quality}, {Stability}, and {Variation} (Feb 2018), \url{http://arxiv.org/abs/1710.10196}, arXiv:1710.10196 [cs, stat]

\bibitem{karras_style-based_2019}
Karras, T., Laine, S., Aila, T.: A style-based generator architecture for generative adversarial networks. IEEE Trans. Pattern Anal. Mach. Intell.  \textbf{43}(12),  4217–4228 (Dec 2021). \doi{10.1109/TPAMI.2020.2970919}, \url{https://doi.org/10.1109/TPAMI.2020.2970919}

\bibitem{karras_analyzing_2020}
Karras, T., Laine, S., Aittala, M., Hellsten, J., Lehtinen, J., Aila, T.: Analyzing and improving the image quality of stylegan. In: Proceedings of the IEEE/CVF conference on computer vision and pattern recognition. pp. 8110--8119 (2020)

\bibitem{kingma_auto-encoding_2013}
Kingma, D.P., Welling, M.: Auto-encoding variational bayes. arXiv preprint arXiv:1312.6114  (2013)

\bibitem{kovar_automated_2004}
Kovar, L., Gleicher, M.: Automated extraction and parameterization of motions in large data sets. ACM Trans. Graph.  \textbf{23}(3),  559–568 (Aug 2004). \doi{10.1145/1015706.1015760}, \url{https://doi.org/10.1145/1015706.1015760}

\bibitem{kovar_motion_2002}
Kovar, L., Gleicher, M., Pighin, F.: Motion graphs. ACM Trans. Graph.  \textbf{21}(3),  473--482 (Jul 2002). \doi{10.1145/566654.566605}, \url{https://doi.org/10.1145/566654.566605}, place: New York, NY, USA Publisher: Association for Computing Machinery

\bibitem{lawrence_gaussian_2003}
Lawrence, N.: Gaussian {Process} {Latent} {Variable} {Models} for {Visualisation} of {High} {Dimensional} {Data}. In: Advances in {Neural} {Information} {Processing} {Systems}. vol.~16. MIT Press (2003), \url{https://proceedings.neurips.cc/paper_files/paper/2003/hash/9657c1fffd38824e5ab0472e022e577e-Abstract.html}

\bibitem{le_music-driven_2023}
Le, N., Pham, T., Do, T., Tjiputra, E., Tran, Q.D., Nguyen, A.: Music-{Driven} {Group} {Choreography}. In: 2023 {IEEE}/{CVF} {Conference} on {Computer} {Vision} and {Pattern} {Recognition} ({CVPR}). pp. 8673--8682. IEEE, Vancouver, BC, Canada (Jun 2023). \doi{10.1109/CVPR52729.2023.00838}, \url{https://ieeexplore.ieee.org/document/10205408/}

\bibitem{lee2022manifolds}
Lee, J.M.: Manifolds and differential geometry, vol.~107. American Mathematical Society (2022)

\bibitem{lee_interactive_2002}
Lee, J., Chai, J., Reitsma, P.S., Hodgins, J.K., Pollard, N.S.: Interactive control of avatars animated with human motion data. In: Proceedings of the 29th annual conference on {Computer} graphics and interactive techniques. pp. 491--500 (2002). \doi{10.1145/566570.566607}

\bibitem{lee_motion_2014}
Lee, Y., Wampler, K., Bernstein, G., Popovi\'{c}, J., Popovi\'{c}, Z.: Motion fields for interactive character locomotion. ACM Trans. Graph.  \textbf{29}(6) (Dec 2010). \doi{10.1145/1882261.1866160}, \url{https://doi.org/10.1145/1882261.1866160}

\bibitem{levine_space-time_2011}
Levine, S., Lee, Y., Koltun, V., Popović, Z.: Space-time planning with parameterized locomotion controllers. ACM Transactions on Graphics  \textbf{30}(3),  1--11 (May 2011). \doi{10.1145/1966394.1966402}, \url{https://dl.acm.org/doi/10.1145/1966394.1966402}

\bibitem{levine_continuous_2012}
Levine, S., Wang, J.M., Haraux, A., Popović, Z., Koltun, V.: Continuous character control with low-dimensional embeddings. ACM Transactions on Graphics  \textbf{31}(4),  1--10 (Aug 2012). \doi{10.1145/2185520.2185524}, \url{https://dl.acm.org/doi/10.1145/2185520.2185524}

\bibitem{li_ai_2021}
Li, R., Yang, S., Ross, D.A., Kanazawa, A.: Ai choreographer: Music conditioned 3d dance generation with aist++. In: Proceedings of the IEEE/CVF International Conference on Computer Vision. pp. 13401--13412 (2021)

\bibitem{liu2019ntu}
Liu, J., Shahroudy, A., Perez, M., Wang, G., Duan, L.Y., Kot, A.C.: Ntu rgb+ d 120: A large-scale benchmark for 3d human activity understanding. IEEE transactions on pattern analysis and machine intelligence  \textbf{42}(10),  2684--2701 (2019)

\bibitem{liu2011markerless}
Liu, Y., Stoll, C., Gall, J., Seidel, H.P., Theobalt, C.: Markerless motion capture of interacting characters using multi-view image segmentation. In: CVPR 2011. pp. 1249--1256. Ieee (2011)

\bibitem{loper_smpl_2023}
Loper, M., Mahmood, N., Romero, J., Pons-Moll, G., Black, M.J.: Smpl: a skinned multi-person linear model. ACM Trans. Graph.  \textbf{34}(6) (Oct 2015). \doi{10.1145/2816795.2818013}, \url{https://doi.org/10.1145/2816795.2818013}

\bibitem{lucas_posegpt_2022}
Lucas, T., Baradel, F., Weinzaepfel, P., Rogez, G.: {PoseGPT}: {Quantization}-based {3D} {Human} {Motion} {Generation} and {Forecasting} (Oct 2022), \url{http://arxiv.org/abs/2210.10542}, arXiv:2210.10542 [cs]

\bibitem{ma20233d}
Ma, X., Su, J., Wang, C., Zhu, W., Wang, Y.: 3d human mesh estimation from virtual markers. In: Proceedings of the IEEE/CVF Conference on Computer Vision and Pattern Recognition. pp. 534--543 (2023)

\bibitem{min_motion_2012}
Min, J., Chai, J.: Motion graphs++: a compact generative model for semantic motion analysis and synthesis. ACM Transactions on Graphics  \textbf{31}(6),  1--12 (Nov 2012). \doi{10.1145/2366145.2366172}, \url{https://dl.acm.org/doi/10.1145/2366145.2366172}

\bibitem{moeslund2006survey}
Moeslund, T.B., Hilton, A., Kr{\"u}ger, V.: A survey of advances in vision-based human motion capture and analysis. Computer vision and image understanding  \textbf{104}(2-3),  90--126 (2006)

\bibitem{osman_star_2020}
Osman, A.A.A., Bolkart, T., Black, M.J.: {STAR}: {Sparse} {Trained} {Articulated} {Human} {Body} {Regressor} (Aug 2020). \doi{10.1007/978-3-030-58539-6_36}, \url{http://arxiv.org/abs/2008.08535}, arXiv:2008.08535 [cs]

\bibitem{pavlakos_expressive_2019}
Pavlakos, G., Choutas, V., Ghorbani, N., Bolkart, T., Osman, A.A., Tzionas, D., Black, M.J.: Expressive {Body} {Capture}: {3D} {Hands}, {Face}, and {Body} {From} a {Single} {Image}. In: 2019 {IEEE}/{CVF} {Conference} on {Computer} {Vision} and {Pattern} {Recognition} ({CVPR}). pp. 10967--10977. IEEE, Long Beach, CA, USA (Jun 2019). \doi{10.1109/CVPR.2019.01123}, \url{https://ieeexplore.ieee.org/document/8953319/}

\bibitem{peng_ase_2022}
Peng, X.B., Guo, Y., Halper, L., Levine, S., Fidler, S.: {ASE}: large-scale reusable adversarial skill embeddings for physically simulated characters. ACM Transactions on Graphics  \textbf{41}(4),  1--17 (Jul 2022). \doi{10.1145/3528223.3530110}, \url{https://dl.acm.org/doi/10.1145/3528223.3530110}

\bibitem{peng_amp_2021}
Peng, X.B., Ma, Z., Abbeel, P., Levine, S., Kanazawa, A.: {AMP}: adversarial motion priors for stylized physics-based character control. ACM Transactions on Graphics  \textbf{40}(4),  1--20 (Aug 2021). \doi{10.1145/3450626.3459670}, \url{https://dl.acm.org/doi/10.1145/3450626.3459670}

\bibitem{petrovich_action-conditioned_2021}
Petrovich, M., Black, M.J., Varol, G.: Action-{Conditioned} {3D} {Human} {Motion} {Synthesis} with {Transformer} {VAE} (Sep 2021), \url{http://arxiv.org/abs/2104.05670}, arXiv:2104.05670 [cs]

\bibitem{petrovich_temos_2022}
Petrovich, M., Black, M.J., Varol, G.: {TEMOS}: {Generating} diverse human motions from textual descriptions (Jul 2022), \url{http://arxiv.org/abs/2204.14109}, arXiv:2204.14109 [cs] version: 2

\bibitem{raab_modi_2023}
Raab, S., Leibovitch, I., Li, P., Aberman, K., Sorkine-Hornung, O., Cohen-Or, D.: Modi: Unconditional motion synthesis from diverse data. In: Proceedings of the IEEE/CVF Conference on Computer Vision and Pattern Recognition. pp. 13873--13883 (2023)

\bibitem{rose_verbs_1998}
Rose, C., Cohen, M., Bodenheimer, B.: Verbs and adverbs: multidimensional motion interpolation. IEEE Computer Graphics and Applications  \textbf{18}(5),  32--40 (Oct 1998). \doi{10.1109/38.708559}, \url{http://ieeexplore.ieee.org/document/708559/}

\bibitem{shafir_human_2023}
Shafir, Y., Tevet, G., Kapon, R., Bermano, A.H.: Human motion diffusion as a generative prior. arXiv preprint arXiv:2303.01418  (2023)

\bibitem{sonderby_ladder_2016}
S\o{}nderby, C.K., Raiko, T., Maal\o{}e, L., S\o{}nderby, S.K., Winther, O.: Ladder variational autoencoders. In: Proceedings of the 30th International Conference on Neural Information Processing Systems. p. 3745–3753. NIPS'16, Curran Associates Inc., Red Hook, NY, USA (2016)

\bibitem{song2020score}
Song, Y., Sohl-Dickstein, J., Kingma, D.P., Kumar, A., Ermon, S., Poole, B.: Score-based generative modeling through stochastic differential equations. arXiv preprint arXiv:2011.13456  (2020)

\bibitem{starke_motion_2023}
Starke, P., Starke, S., Komura, T., Steinicke, F.: Motion {In}-{Betweening} with {Phase} {Manifolds}. Proceedings of the ACM on Computer Graphics and Interactive Techniques  \textbf{6}(3),  1--17 (Aug 2023). \doi{10.1145/3606921}, \url{https://dl.acm.org/doi/10.1145/3606921}

\bibitem{starke_deepphase_2022}
Starke, S., Mason, I., Komura, T.: {DeepPhase}: periodic autoencoders for learning motion phase manifolds. ACM Transactions on Graphics  \textbf{41}(4),  1--13 (Jul 2022). \doi{10.1145/3528223.3530178}, \url{https://dl.acm.org/doi/10.1145/3528223.3530178}

\bibitem{starke_neural_2021}
Starke, S., Zhao, Y., Zinno, F., Komura, T.: Neural animation layering for synthesizing martial arts movements. ACM Transactions on Graphics  \textbf{40}(4),  1--16 (Aug 2021). \doi{10.1145/3450626.3459881}, \url{https://dl.acm.org/doi/10.1145/3450626.3459881}

\bibitem{sutton1998reinforcement}
Sutton, R.S., Barto, A.G.: The reinforcement learning problem. Reinforcement learning: An introduction pp. 51--85 (1998)

\bibitem{tang_dance_2018}
Tang, T., Jia, J., Mao, H.: Dance with {Melody}: {An} {LSTM}-autoencoder {Approach} to {Music}-oriented {Dance} {Synthesis}. In: Proceedings of the 26th {ACM} international conference on {Multimedia}. pp. 1598--1606. ACM, Seoul Republic of Korea (Oct 2018). \doi{10.1145/3240508.3240526}, \url{https://dl.acm.org/doi/10.1145/3240508.3240526}

\bibitem{tessler_calm_2023}
Tessler, C., Kasten, Y., Guo, Y., Mannor, S., Chechik, G., Peng, X.B.: {CALM}: {Conditional} {Adversarial} {Latent} {Models} for {Directable} {Virtual} {Characters}. In: Special {Interest} {Group} on {Computer} {Graphics} and {Interactive} {Techniques} {Conference} {Conference} {Proceedings}. pp.~1--9. ACM, Los Angeles CA USA (Jul 2023). \doi{10.1145/3588432.3591541}, \url{https://dl.acm.org/doi/10.1145/3588432.3591541}

\bibitem{tevet_human_2022}
Tevet, G., Raab, S., Gordon, B., Shafir, Y., Cohen-Or, D., Bermano, A.H.: Human {Motion} {Diffusion} {Model} (Oct 2022), \url{http://arxiv.org/abs/2209.14916}, arXiv:2209.14916 [cs]

\bibitem{tu2011manifolds}
Tu, L.W.: Manifolds. In: An Introduction to Manifolds, pp. 47--83. Springer (2011)

\bibitem{noauthor_neural_2017}
Van Den~Oord, A., Vinyals, O., et~al.: Neural {Discrete} {Representation} {Learning}. Advances in neural information processing systems  \textbf{30} (2017)

\bibitem{wang_spatio-temporal_2019}
Wang, H., Ho, E.S.L., Shum, H.P.H., Zhu, Z.: Spatio-temporal {Manifold} {Learning} for {Human} {Motions} via {Long}-horizon {Modeling} (Aug 2019), \url{http://arxiv.org/abs/1908.07214}, arXiv:1908.07214 [cs]

\bibitem{xu_ghum_2020}
Xu, H., Bazavan, E.G., Zanfir, A., Freeman, W.T., Sukthankar, R., Sminchisescu, C.: {GHUM} \& {GHUML}: {Generative} {3D} {Human} {Shape} and {Articulated} {Pose} {Models}. In: 2020 {IEEE}/{CVF} {Conference} on {Computer} {Vision} and {Pattern} {Recognition} ({CVPR}). pp. 6183--6192. IEEE, Seattle, WA, USA (Jun 2020). \doi{10.1109/CVPR42600.2020.00622}, \url{https://ieeexplore.ieee.org/document/9157563/}

\bibitem{yu_structure-aware_2020}
Yu, P., Zhao, Y., Li, C., Yuan, J., Chen, C.: Structure-{Aware} {Human}-{Action} {Generation} (Aug 2020), \url{http://arxiv.org/abs/2007.01971}, arXiv:2007.01971 [cs, stat]

\bibitem{zeng_human_2021}
Zeng, R., Dai, J., Bai, J., Pan, J., Qin, H.: Human motion synthesis and control via contextual manifold embedding. In: PG (Short Papers, Posters, and Work-in-Progress Papers). pp. 25--30 (2021)

\bibitem{zhang_motiondiuse_2024}
Zhang, M., Cai, Z., Pan, L., Hong, F., Guo, X., Yang, L., Liu, Z.: Motiondiffuse: Text-driven human motion generation with diffusion model. IEEE Trans. Pattern Anal. Mach. Intell.  \textbf{46}(6),  4115–4128 (Jan 2024). \doi{10.1109/TPAMI.2024.3355414}, \url{https://doi.org/10.1109/TPAMI.2024.3355414}

\bibitem{zhang_remodiffuse_2023}
Zhang, M., Guo, X., Pan, L., Cai, Z., Hong, F., Li, H., Yang, L., Liu, Z.: {ReMoDiffuse}: {Retrieval}-{Augmented} {Motion} {Diffusion} {Model} (Apr 2023), \url{http://arxiv.org/abs/2304.01116}, arXiv:2304.01116 [cs]

\bibitem{zhang_place_2020}
Zhang, S., Zhang, Y., Ma, Q., Black, M.J., Tang, S.: {PLACE}: {Proximity} {Learning} of {Articulation} and {Contact} in {3D} {Environments} (Nov 2020), \url{http://arxiv.org/abs/2008.05570}, arXiv:2008.05570 [cs]

\bibitem{zhang2021we}
Zhang, Y., Black, M.J., Tang, S.: We are more than our joints: Predicting how 3d bodies move. In: Proceedings of the IEEE/CVF Conference on Computer Vision and Pattern Recognition. pp. 3372--3382 (2021)

\bibitem{zhu_motionbert_2023}
Zhu, W., Ma, X., Liu, Z., Liu, L., Wu, W., Wang, Y.: {MotionBERT}: {A} {Unified} {Perspective} on {Learning} {Human} {Motion} {Representations} (Aug 2023), \url{http://arxiv.org/abs/2210.06551}, arXiv:2210.06551 [cs]

\bibitem{zhu_human_2023}
Zhu, W., Ma, X., Ro, D., Ci, H., Zhang, J., Shi, J., Gao, F., Tian, Q., Wang, Y.: Human {Motion} {Generation}: {A} {Survey} (Nov 2023), \url{http://arxiv.org/abs/2307.10894}, arXiv:2307.10894 [cs]

\bibitem{zuo2024loose}
Zuo, C., Wang, Y., Zhan, L., Guo, S., Yi, X., Xu, F., Qin, Y.: Loose inertial poser: Motion capture with imu-attached loose-wear jacket. In: Proceedings of the IEEE/CVF Conference on Computer Vision and Pattern Recognition. pp. 2209--2219 (2024)

\end{thebibliography}

\end{document}